# A Deep Learning-Based Approach for Measuring the Domain Similarity of Persian Texts

**Hossein Keshavarz**

Department of Computer Engineering
Sharif University of Technology
hkeshavarz@ce.sharif.edu

**Shohreh Tabatabayi Seifi**

Department of Computer Engineering
Sharif University of Technology
tabatabayiseifi@ce.sharif.edu

**Mohammad Izadi**

Department of Computer Engineering
Sharif University of Technology
izadi@sharif.edu

## Abstract

In this paper, we propose a novel approach for measuring the degree of similarity between categories of two pieces of Persian text, which were published as descriptions of two separate advertisements. We built an appropriate dataset for this work using a dataset which consists of advertisements posted on an e-commerce website. We generated a significant number of paired texts from this dataset and assigned each pair a score from 0 to 3, which demonstrates the degree of similarity between the domains of the pair. In this work, we represent words with word embedding vectors derived from word2vec. Then deep neural network models are used to represent texts. Eventually, we employ concatenation of absolute difference and bit-wise multiplication and a fully-connected neural network to produce a probability distribution vector for the score of the pairs. Through a supervised learning approach, we trained our model on a GPU, and our best model achieved an $F_1$ score of 0.9865.

## 1. Introduction

One of the most challenging problems in natural language processing (NLP) is to figure out whether two texts belong to the same domain or not. This problem has applications in vast areas such as smart advertising robots, which look for semantically related pages to show an ad, or robots, which are supposed to collect news pertinent to a specific topic. Generally speaking, a measure for deciding how close domains of two texts are can be used as a metric for text distance in text classification tasks.

In this research, we strived to develop a system for a task which is a combination of two famous tasks in NLP, namely semantic textual similarity (STS) and text classification. In this task, two texts are compared to each other like the STS task, but rather than their semantics, their classes are the basis of this comparison.

Recently, Cafebazaar Research Group [1] published a dataset which consists of advertisements that were posted on Divar [2] application. We developed a proper dataset to train our proposed model for the mentioned task by the use of these advertisements which are categorized in three levels. The generated dataset includes multitudinous of ad pairs each of which is assigned a score identifying to what extent, categories of advertisements constituting that pair are domain-wise similar. It is unlike the preceding approaches in which the categories of two texts were spotted in the first place, and then, their labels were compared. In our model, the domain similarity of two pieces of text is determined directly.

No one can cast a shadow of doubt on the fact that achieving a considerable result in this problem demands that the computer system precisely understands both input texts; thus an accurate method for representing the semantics of texts must be picked. Significant improvement in computational linguistics in recent years, which was led by utilizing deep neural networks (DNNs) in building models stimulated us to use such networks to implement our system. Particularly, we trained our model via convolutional neural networks (CNNs) and long short-term memory networks (LSTMs) separately. The former is vastly used for classification problems in both NLP and image processing fields, and the latter one captures and maintains long-term dependencies. This fact causes LSTMs to be able to effectively represent features of texts, which is the preliminary to obtain extraordinary results in NLP problems.

It is worth mentioning eliciting accurate features for texts entails proper word representation. To achieve this, we made use of word embedding. The processing resource which facilitated the computation for making word embedding vectors and training the whole model was an Nvidia GeForce GTX 1080 Ti graphical processing unit (GPU).

This article is organized as follows: Section 2 presents the related works, in Section 3 the dataset is introduced, in Section 4 our method is described, Section 5 demonstrates the experimental results and then in Section 6 we discuss future possible works and the conclusion.

## 2. Related works

Before the advent of deep learning methods in solving NLP problems, the solutions in this field of study were limited to rule-based and statistical methods. Achieving noticeable results in computer vision tasks with the use of deep neural networks propelled NLP researchers to attempt exploiting these networks in their systems.

Bengio et al. [3] developed a neural network-based system to learn distributed representations for words which embody the probability for word sequences. This work became the basis for further language modeling research and led to wider use of distributed representations for words known as word embedding. Word embedding captures semantical, syntactical and lexical similarities of words within a corpus. Word embedding is a great way to solve the sparsity problem of data and maps high dimensional features of text to lower dimensional representations containing useful features. Obtaining low dimensional representations of inputs that encompass critical information paves the way for achieving a great result in further evaluation [4]. In recent years, a great number of effective approaches for word embedding proposed [5-8]. Word2vec [6] has been proved to be a competent method for learning distributed features of texts, and it was utilized in this work.

In addition to learning features of words, deep neural networks also were exploited for sentence representations. Convolutional neural networks (CNNs) and recurrent neural networks (RNNs) [9] are two primary architectures which are extensively used to handle tasks in natural language processing.

CNNs were initially used in computer vision and image processing. CNNs take windows of features, apply filters on local features, and convolve these filters to capture the most activated features. The success of CNNs in image processing became the impetus for NLP researchers to employ these networks for feature extraction. Collobert et al. [10] and Collobert et al. [5] utilized CNNs for traditional NLP problems like part-of-speech tagging and named entity recognition in which given a sentence, words were supposed to be assigned labels based on their semantic roles. In the architectures of these two works, windows of words or the whole sentence were chosen, and the role of words within them were tagged based on the features a CNN extracted followed by classical neural networks to generate probability distribution. In another work, Nguyen et al. [11] train a CNN on top of a pre-trained word embedding for relation extraction task. The convolutional layer in this work is followed by a max pooling layer for downsampling.

On the other hand, RNNs are mighty architectures for capturing long range dependencies. RNNs are subject to gradient exploding or gradient vanishing. LSTM [12] is a variant of RNNs that solve this problem. As these networks maintain earlier information, they are a good option for text processing. It has been widely utilized for sentence modeling. Liu et al. [13] introduce an approach based on multi-timescale LSTM to model long sentences efficiently and avoid missing very long range relations. Sutskever et al. [14] build an encoder followed by a decoder with LSTM for machine translation. Numerous works propose a hybrid method in which their architecture is made up of both CNNs and LSTMs for sentence embedding in various NLP tasks [15-17].

Generally, CNNs enjoy hierarchical structures. This fact causes them to be capable of extracting key phrases inside the sentences; thus, they are superior in classification problems [18]. On the contrary, if the task is a sort of sequence modeling problem, variants of RNN are more appropriate choices [19]. However, this is not a definite fact, and there exist copious exceptions [20].

Kalchbrenner et al. [21] propose a CNN architecture with a dynamic max pooling layer for sentence modeling. Its results on classification tasks are notable. Some other works build their model on top of pre-trained word embeddings to achieve surpassing results [22]. Kim [22] reports that although static word embedding weights achieves astounding results, enabling the system to fine-tune these weights outperforms the static mode.

Lai et al. [23] and Liu et al. [24] train RNNs on top of word2vec pre-trained word embedding. The former utilizes bidirectional RNN followed by max pooling layer, and the latter develops a multi-tasks approach in order to have a sufficient number of samples which is a necessity for deep learning methods. They both use Softmax function to classify multi-class texts.

As we stated earlier, the task we are coping with in this work is a combination of text classification problem and semantic textual similarity (STS) problem because it involves a direct comparison of two sentences. STS task became increasingly popular in recent years mostly because of SemEval competition. SemEval incorporated this task from 2012 until 2017 [25-30]. Lots of deep learning methods propose to handle this task. Two major neural networks that were exploited in such methods were CNNs and LSTMs. He et al. [31] model two input sentences with CNNs with Siamese structure [32] in which two CNNs share their weights, and they are trained in parallel. Shao [33] trains two CNNs for two input sentences on top of pre-trained GloVe word vectors. CNNs are followed by max pooling layers. It ranked 3rd on the primary track of SemEval 2017. Sanborn et al. [34] set some experiments to compare the performance of RNN extension for language modelling [35] with recursive neural networks [36]. This work does not focus on a specific task, and its purpose is developing a universal model for semantic similarity.

Although our task is similar to STS, there exist some fundamental differences. Concretely, two sentences can be of the same class, but the amount of information in one of them surpasses the amount of information in the other one. In this case, in our task, these sentences count similar. On the other hand, they do not receive the best score in STS task. In this respect, the task herein is much like the works in [37] and [38] in which general similarity is measured with CNNs.

Unfortunately, unlike languages like English, there has been no considerable research conducted on Persian. Persian writing system is a challenging one. It shares a lot of characteristics with Arabic, but a few works are carried out on Arabic too. [39] is one the few works on text classification with deep learning methods on Arabic in which CNNs are utilized to capture semantic information of texts.

## 3. Data Collection

Cafebazaar Research Group recently published a dataset for research purposes. There are 947635 records of the posted advertisements on Divar application in this dataset. Divar application is a classified e-commerce platform in which people are able to directly sell their products or introduce their services by posting ads containing information about their products and services. A random sample record is shown in table 1.

As it is shown in figure 1, all of the advertisements are categorized in 3 levels. The "cat1" column demonstrates the categories of advertisements at a high level, and through "cat2" and "cat3", the classes of ads become more precise. In the entire dataset, 87 distinct combinations of three levels of categories exist. There are 13 more attributes than categories columns in the dataset two of which are of more importance to us, namely description ("desc") and title.

| archive_by_user | False |
|---|---|
| brand | Nokia:: نوکیا[1] |
| cat1 | electronic-devices |
| cat2 | mobile-tablet |
| cat3 | mobile-phone |
| city | Qom |
| created_at | Wednesday 07AM |
| desc[2] | سلام.یه گوشیه6303سالم که فقط دوتا خط کوچیک رو ال سی دیشه با شارژر فابریک |
| id | 36762902012926 |
| image_count | 2 |
| mileage | - |
| platform | mobile |
| price | 60000 |
| title | نوکیا6303[3] |
| type | - |
| year | - |

*Table 1. A sample record in the dataset of Divar*

With the description of our task in mind, it is obvious that we were not able to employ this rich dataset in its original form because we needed pair texts to assess the degree of their categories' similarity. In order to cross the aforementioned barrier, we had to modify the dataset and construct the desirable dataset for our intention in following steps:

- For each record, we joined the description of ad to the title of that ad. These two columns are made up of the text people usually add to their advertisements in order to describe the product they wish to purchase or sell. In the first place, it may seem unnecessary to use titles in addition to descriptions because most of the advertisements possessed descriptions written in great detail, but we observed that there is also an abundance of ads whose descriptions were utterly dependent on their titles. For instance, a lot of the ads in the mobile phone category were posted in the way that their ad publishers mentioned the brand of their phones in the title, and then in the description section, technical or outward characteristics of the cell phones were provided. For these

---
1. English translation: Nokia
2. English translation: "Hi. It's a 6303 cell-phone which only have two scratches on its screen, and it comes with an original battery charger."
3. English Translation: Nokia 6303

- advertisements, the result of joining the descriptions to the titles produced more meaningful texts while this decision did not adversely affect the ads with comprehensive descriptions.
- We defined the constraint that texts comprising each pair must not be inordinately different in terms of their length. Then, one at a time, we selected two texts which met our defined constraint in a haphazard manner and made them pair with each other. Concretely, for making every single pair, we picked two texts randomly and checked that the length difference of these two texts was not more than 15 percent of the length of the shorter text after rounding. If it was, we would ignore the pair; otherwise, we collected the pair and repeated the mentioned procedure to create the next pair.
- We concatenated "cat1", "cat2", "cat3" attributes in sequence to create category combinations. Then, each text was bound with its corresponding combination. Subsequently, every pair was mapped to the score of the similarity between texts building that pair. This score was measured based on the combinations of the classes of two texts within the pair. We established a quite straightforward approach through which we compared the combinations of the categories of two texts. When they were completely the same, we gave the pair score 3, when they were varied in their last level of category ("cat3"), they were assigned score 2, when they were different in their last two levels of category ("cat2" and "cat3"), they were given score 1 and eventually, when they were completely different, the score of the pair was 0.

In the process described above, we created 20 million pair texts to use as the dataset for training our networks on. For the sake of having a uniform distribution, we became sure that each class of scores had 5 million pairs. To achieve this, we just controlled the number of text pairs in each class. We did not manipulate the method of picking texts to maintain the random manner.

A pivotal point which deserves some words here is that the original Divar dataset includes a substantial number of Persian sentences in informal and colloquial form. There are also lots of misspelling, grammatical mistakes and punctuation errors due to the fact that various sorts of people with different professions and educational backgrounds wrote these texts; thus, the ultimate dataset derives these characteristics, which escalates the complication of a precise text processing, from the original one.

Moreover, another challenge arising here is that since the context of prepared texts is advertising, there is a myriad of cases in which the category combinations of texts in a pair match although they encompass fundamentally different entities within them. For the purpose of illustration, consider a pair made up of two advertisements whose categories are "light car." Although this pair must receive the highest score (score 3), it is likely that the two texts include information about completely different car brands and models.

## 4. Description of the method

Figure 1 illustrates the workflow of our model. In the first place, the word embedding vectors were built, and 2-D text vectors were made up of them. Then we exploited deep neural networks to capture compressed feature vectors of texts. The elicited vectors of texts in each pair were compared, and subsequently, a fully connected neural network was employed to produce probability distributions of the score of similarity. In the following, each phase will be elucidated in more detail.

### 4.1. Word Representation

Methods like term-frequency (tf), term-frequency inverse-document-frequency (tf-idf), bag of words (BoW) and word co-occurrence are general statistics, which extensively were used in characterizing words in given documents before neural networks demonstrate their power in natural language processing. The key problem of these approaches is that the generated features of the text on which these methods are applied are extremely sparse and consequently, redundant across many dimensions. Their dimensionality also is needed to be reduced via other approaches like PCA.

Another popular approach is mapping words in documents to continuous vectors of real numbers with low dimensions. This approach is known as word embedding. In recent years, neural learning algorithms proved to be beneficial in creating word representations in this way, and research has shown that the produced vectors in this manner encompass semantic features of words effectively if colossal corpora are fed to the models. As one of the most decisive prerequisites for obtaining significant results in deep neural network-based tasks is having appropriate representation vectors for words, we acquired a massive Persian Corpus to create word representations via word embedding method. We also conducted a series of preprocessing activities, and then, the word embedding vectors were built. In the following subsections, these three stages are explained.

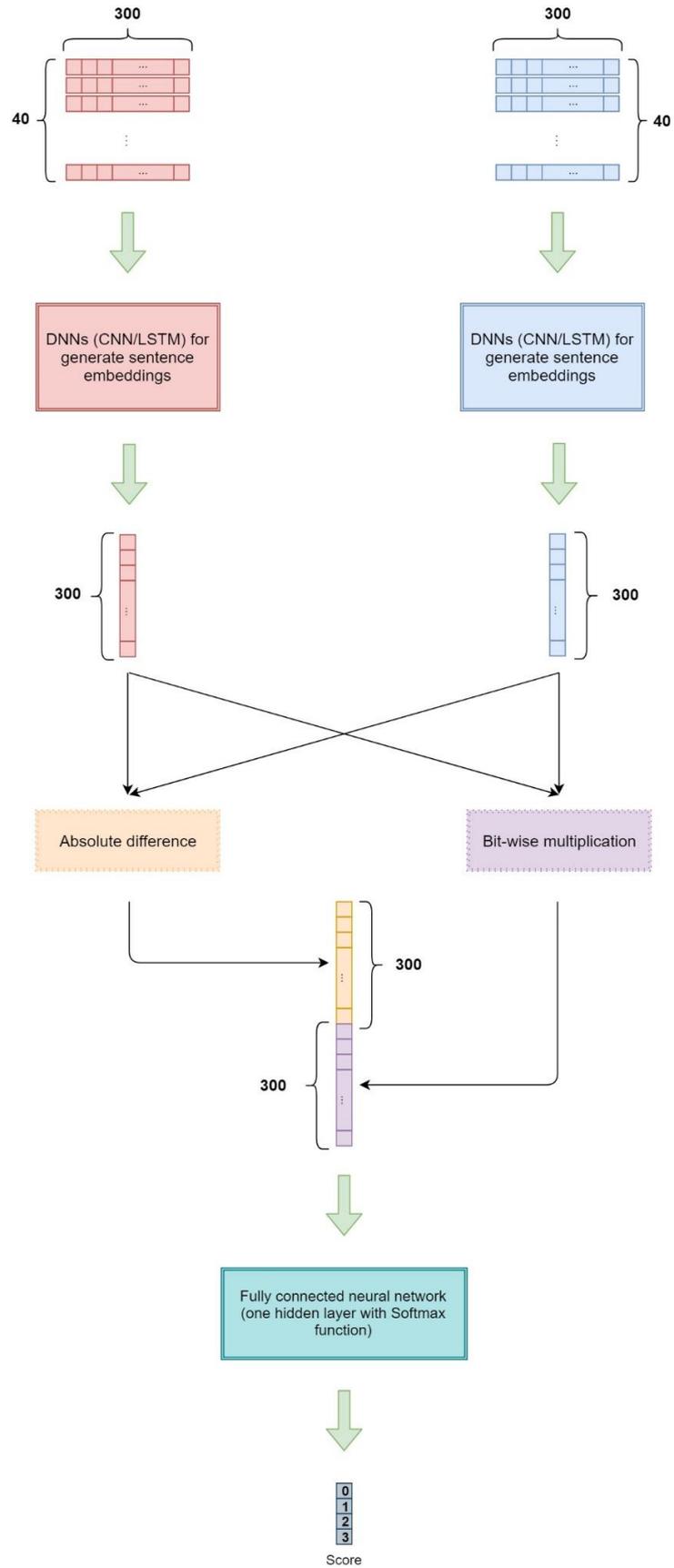
*Figure 1. An overview of the workflow of system*

### 4.1.1. Corpus

Unfortunately, there is no proper pre-trained model in Persian to meet requirements of the defined task since most of the available models are based on the articles reside in Persian Wikipedia [40] whose styles of writing are mostly scientific. Since ordinary people wrote the descriptions of the advertisements, and these texts are often in colloquial form, such models cannot correctly encompass the information needed for further evaluations. To address this problem, we decided to build word vectors via word embedding method by ourselves; thus, we needed a comprehensive corpus to train. We acquired a Persian corpus called "Peykare," which contains 100 million words [41]. This corpus is the most comprehensive Persian corpus. We added advertisement texts to this corpus in order to have a monumental corpus for creating word embedding vectors. Moreover, we contend that in this way, ultimate word vectors incorporate advertising semantics within the texts in the dataset.

### 4.1.2. Preprocessing

As it was mentioned earlier that the texts of the dataset were provided by people from different social classes with varied education backgrounds, an adequate and appropriate approach for preprocessing is of paramount importance; therefore, we combined several distinct methods to prepare the dataset. First of all, a myriad of exotic symbols was used in the texts; hence, we eliminated ones that were more common among these symbols. It must be taken into account that elimination is not the best idea, and in some cases, if it was possible, we altered the symbols to proper ones. For example, in several cases, features of products were separated by "*" symbol instead of "." character. In addition, "." characters were not followed by a white space; thus, tokenizing the texts was subject to producing erroneous tokens. Hence, we replaced all such characters with space characters. After this process, we used Hazm [42], which is a library for cleaning Persian text, to normalize the texts. During this process, Arabic specific characters like "ي" and "ك" were mapped to their equivalent Persian Unicode (in this example "ی" and "ک"), and variant forms of some Persian characters were unified to a single Unicode representation (for the purpose of illustration, "گلها", "گل ها" and "گلهٰا" are three different representations of the same word meaning "flowers" in Persian). Eventually, we prepared a list of 330 most frequent stop words in Persian and removed these words if they existed within the texts. A number of these words are shown in table 2. In order to maintain uniformity, we preprocessed the texts in Peykare corpus in the same way.

| Persian stop words | | | | | | | | | | | | | | | |
|---|---|---|---|---|---|---|---|---|---|---|---|---|---|---|---|
| و | در | به | از | است | برای | یک | ها | شود | تا | هستند | برخی | شاید | خواهد | با | ...* |

*Table 2. Some Persian stop words*

* English counterparts from left to right: "and", "in", "to", "from", "is", "for", "a", plural "s", "be", "to" / "until", "are", "some", "maybe", "will", "with", …

### 4.1.3. Word Embedding

To develop word embedding vectors, we deployed word2vec method in continuous bag of words (CBOW) mode [6] owing to the experiments which were mentioned in [43] and carried out on Persian text. According to this research, this mode of word2vec brings about superior results in experiments on Persian corpora. We trained word2vec on the combination of ad texts and Peykare corpus after they were preprocessed. We set the lower limit of word frequencies to 2 in order to exclude rare words from the words that we were supposed to create word embedding vectors for. After the training process, we were provided with 597897 vectors, and the dimension of each vector was 300. In addition, we enabled this layer to be fine-tuned during the training of the model to achieve better results [22].

### 4.2. Deep Neural Network Layer

Up to this point, we have word embedding vectors; therefore, we can develop 2-dimensional matrices for texts. The only issue was that the texts in the dataset were not of the same length; hence, we set the length of 40 words as the maximum length of texts to become capable of processing fixed-length text matrices. The texts with more words were trimmed on their end. For texts with fewer words than this fixed length, we developed a special sort of padding in which we added the words of the same text from its beginning to the end until the length of that text reaches to 40 words. In most previous works, the padding approach was adding zero vectors, but in our case, because of the nature of the dataset, there were lots of rare words in the corpus for which we did not create word embedding vectors as we mentioned in the previous subsection. Therefore, zero padding could lead to information miss, and our experiments proved this hypothesis.

The ultimate purpose of the whole architecture we proposed is acquiring a quantitative measure for each text so that recognizing the degree of similarity between two texts becomes possible. Thus, we needed to find an avenue to represent every text with a vector. As a result, we utilized deep neural networks (DNNs) for training on texts using the embedding vectors of words and generate vectors that embody the cardinal semantical information of texts effectively.

Effectiveness is a momentous factor in this context because we already have embedding vectors for words, and a naïve approach would be using 40 × 300 matrices of each text. At first glance, this approach seems promising because word representation vectors made by word2vec model proved to be useful in most experiments owing to a large amount of semantic information they contain, but further evaluation of this approach suggests that this method is not effective due to the demanding resources it takes and the sparseness of text matrices. Moreover, it must not be overlooked that in such representation, words are weighted similarly. It is not desirable because most of the words in ad texts do not carry meaningful information. With this piece of information in mind, we clearly need to properly compress these text matrices to obtain less sparse text vectors.

A significant point arising here is that the method which we exploit to do this compression must not neglect important features of texts. DNNs have shown mighty models which correctly extract significant features for this purpose. In this stage, we conducted our experiment with two different types of DNNs.

### 4.2.1. CNN

In the first experiment, we placed two convolutional neural networks (CNNs) for each of the texts which are entered the system as inputs. Like [22] in NLP and [44] in image processing, each CNN was followed by a max pooling layer instead of having bigger strides. A CNN produces feature mappings by filtering every single part of data. The aforementioned mappings indicate how much the corresponding part activated the filter. Perhaps the most significant merit of such networks is that they can be trained very quickly. Nevertheless, previous works indicate that they are staggering models for the sentence embedding usage because these networks are powerful networks in feature extraction. In addition, they do not bias through the learning process.

In this stage, we planted two one-layered CNNs with 3 × 1 filters, "ReLU" [45] activation function and "He Uniform" as kernel weights initializer. We fed these two networks with two texts in each pair we had made independently. Specifically, we represented each text, which was made up of 40 words, using the embedding vectors of its words so that each CNN network was given a 40 × 300 matrix as its input.

But a CNN just convolves data in given windows and cannot compress data on its own, and the outputs of these CNNs are matrices which are the same as the input matrices in terms of dimensions. To fulfill the condensing purpose which we were looking for, CNNs must be followed by max pooling layers. The outputs of CNNs are 38 × 300 matrices owing to the size of filters (the first and the last row of text matrices were ignored). When we apply 38 × 1 max pool on the outputs of CNNs, we are provided with 300-dimensional vectors which consist of most activated data after having the texts convolved. Hence, these 300-dimensional vectors generated by max pooling layer include the most important features of the texts and will be used in the following stages as representations of texts.

### 4.2.2. LSTM

Another DNNs we employed to create text representation vectors from input matrices was LSTM networks. LSTMs are a special kind of Recurrent Neural Networks (RNNs), which are networks with loops in them to capture and maintain information in future iterations. RNNs suffer from not being able to include long-term dependencies in the process of leaning. They are prone to gradient vanishing and gradient explosion. LSTMs, on the other hand, address this issue by using a gated structure and are capable of remembering information for long periods of time; thus, they proved to be helpful in most of the problems pertinent to natural language processing field of study because they make it possible to keep the context of text, which is decisive in future occurrences of words.

Although their advantages, LSTMs are biased toward the data they have recently observed. Hence, they are not as good models as CNNs in feature extraction. Furthermore, learning via an LSTM network is an extremely time-consuming process. However, by using LSTMs in our experiments, the ultimate result was superior to the result of the experiments which we conducted with the CNN-based system.

We exploited two LSTMs with "Tanh" activation function, "Hard Sigmoid" as the activation function to use in the recurrent steps, "Glorot Uniform" as kernel weights matrix initializer and "Orthogonal" for initializing recurrent kernel weights matrix. We fed these two networks with inputs in the same shape that described in CNN section. After LSTM layers work on two input texts, they elicit key characteristics of texts and represent those in two 300-dimensional vectors. By having these two 300-dimensional vectors which represent input texts, the process of assessing the semantical similarity of two texts becomes much more streamlined.

### 4.3. Similarity Comparison

To find how much two generated vectors, which we call semantic vectors, are similar, we exploit the difference measure proposed in [33] and [19] (without including the exact vectors of texts) based on which a 600-dimensional vector will be produced by concatenating absolute difference of two vectors and bitwise-multiplication of them. This 600-dimensional vector determines the degree of similarity between semantic vectors.

### 4.4. FCNN and Probability Distribution

As mentioned in the previous section, the similarity score vector has 600 dimensions, but the ultimate vector we desire to have should only contain four dimensions for score 0 to 3; therefore, we need to compress the similarity score vector and transfer it to a probability distribution. To accomplish this objective, we put a fully-connected neural network (FCNN) comprising one hidden layer with four units. The hidden layer enjoyed "Softmax" activation function which was responsible for manufacturing the intended distribution vector possessed.

### 5. Experiments and Results

We divided the 20 million prepared pairs in the data collection phase into three groups. We randomly selected 10 percent of the whole dataset for testing, 10 percent of the remaining pairs for validation and the rest for training. We utilized validation data to tune parameters. We achieved the best result of our system when the number of epochs was 20 for the CNN-based model (we will discuss the LSTM-based model later), the batch size was 1000, "Adam Optimizer" was hired as the optimization algorithm and "Categorical Cross-entropy" was set as the loss function. Other parameters were introduced in the previous section. We compared $F_1$ score of outputs of the system in each case for validation. Table 3 illustrates the results of our experiments when the network was trained on 16 million text pairs and tested on 2 million text pairs.

As we mentioned in section 4.2, CNNs are much faster than LSTMs. As time is a cardinal factor, we chose it as our criterion; thus, the system with CNN was trained 20 epochs. In contrast, we trained the LSTM-based model only 5 epochs, which took almost the same time as the 20-epoch training of the CNN-based system took. We carried out both experiments on a remote GPU server. The results suggest a slight difference between the effectiveness of these two networks, and as it is shown in table 3, the LSTM-based approach works better even with less training epochs.

Since it was the first research on the introduced dataset, for the sake of comparison, we also conducted another experiment in which the text representation vectors were made by getting the average of the word vectors comprising them [23, 34]. In other words:

$$s = w_1 w_2 \dots w_{40}$$
$$R_{w_i} = [r_{1_i}, r_{2_i}, \dots, r_{300_i}]$$
$$R_s = \left[ \frac{\sum_{i=1}^{40} r_{1_i}}{40}, \frac{\sum_{i=1}^{40} r_{2_i}}{40}, \dots, \frac{\sum_{i=1}^{40} r_{300_i}}{40} \right]$$

In this formula, $s$ is a sample text, $w_i$ is the $i$-th word in $s$, $R_{w_i}$ is the word embedding vector of $w_i$ and $R_s$ is the representation vector of $s$. In this experiment, we generated $R_s$ for each of the two input texts and fed them to the similarity comparison stage followed by FCNN layer just as we discussed in subsections 4.3 and 4.4.

Table 4 shows the results of the proposed systems with respect to the number of training samples on which those systems were trained on. Figure 2 and figure 3 demonstrate the effect of the number of training samples in our experiments on the ultimate result. It clearly indicates that this factor plays an imperative role in our task. The major difference between the score and accuracy of the LSTM-based system and the other two in experiments with fewer training samples is another point that must be taken into account.

| method | Weighted $F_1$ Score | Score 0 $F_1$ Score | Score 1 $F_1$ Score | Score 2 $F_1$ Score | Score 3 $F_1$ Score | Accuracy |
|---|---|---|---|---|---|---|
| word2vec mean | 0.9420 | 0.9494 | 0.9350 | 0.9410 | 0.9427 | 0.9421 |
| CNN-based | 0.9710 | 0.9765 | 0.9716 | 0.9670 | 0.9689 | 0.9710 |
| LSTM-based | **0.9865** | **0.9914** | **0.9900** | **0.9817** | **0.9831** | **0.9865** |

*Table 3. Results*

| Training Sample Number | word2vec mean $F_1$ Score | word2vec mean Accuracy | CNN-based $F_1$ Score | CNN-based Accuracy | LSTM-based $F_1$ Score | LSTM-based Accuracy |
|---|---|---|---|---|---|---|
| 10k | 0.5428 | 0.5428 | 0.5359 | 0.5360 | **0.7472** | **0.7480** |
| 20k | 0.6273 | 0.6290 | 0.6387 | 0.6384 | **0.8117** | **0.8120** |
| 50k | 0.7156 | 0.7134 | 0.7576 | 0.7574 | **0.8598** | **0.8598** |
| 100k | 0.7569 | 0.7568 | 0.8068 | 0.8070 | **0.8907** | **0.8909** |
| 200k | 0.7977 | 0.7980 | 0.8511 | 0.8511 | **0.9110** | **0.9111** |
| 500k | 0.8419 | 0.8419 | 0.8802 | 0.8805 | **0.9310** | **0.9310** |
| 1m | 0.8643 | 0.8646 | 0.9016 | 0.9018 | **0.9448** | **0.9449** |
| 2m | 0.8901 | 0.8901 | 0.9208 | 0.9207 | **0.9588** | **0.9588** |
| 5m | 0.9194 | 0.9194 | 0.9435 | 0.9436 | **0.9747** | **0.9747** |
| 10m | 0.9366 | 0.9367 | 0.9584 | 0.9583 | **0.9832** | **0.9832** |
| 16m | 0.9420 | 0.9421 | 0.9710 | 0.9710 | **0.9865** | **0.9865** |

*Table 4. Performances of systems with respect to the number of training samples*

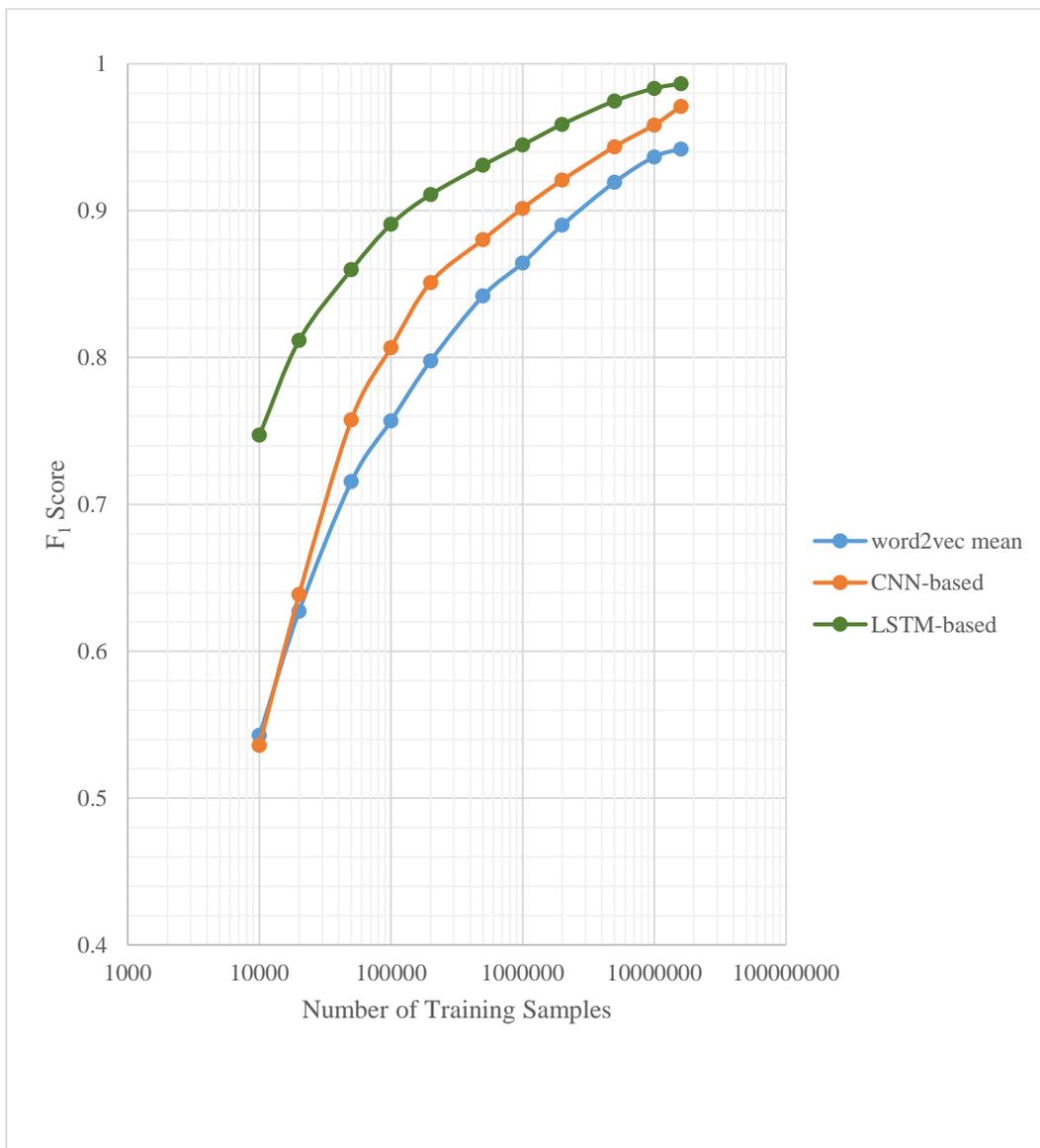

*Figure 2. The effect of the number of the number of training samples on F1 scores*

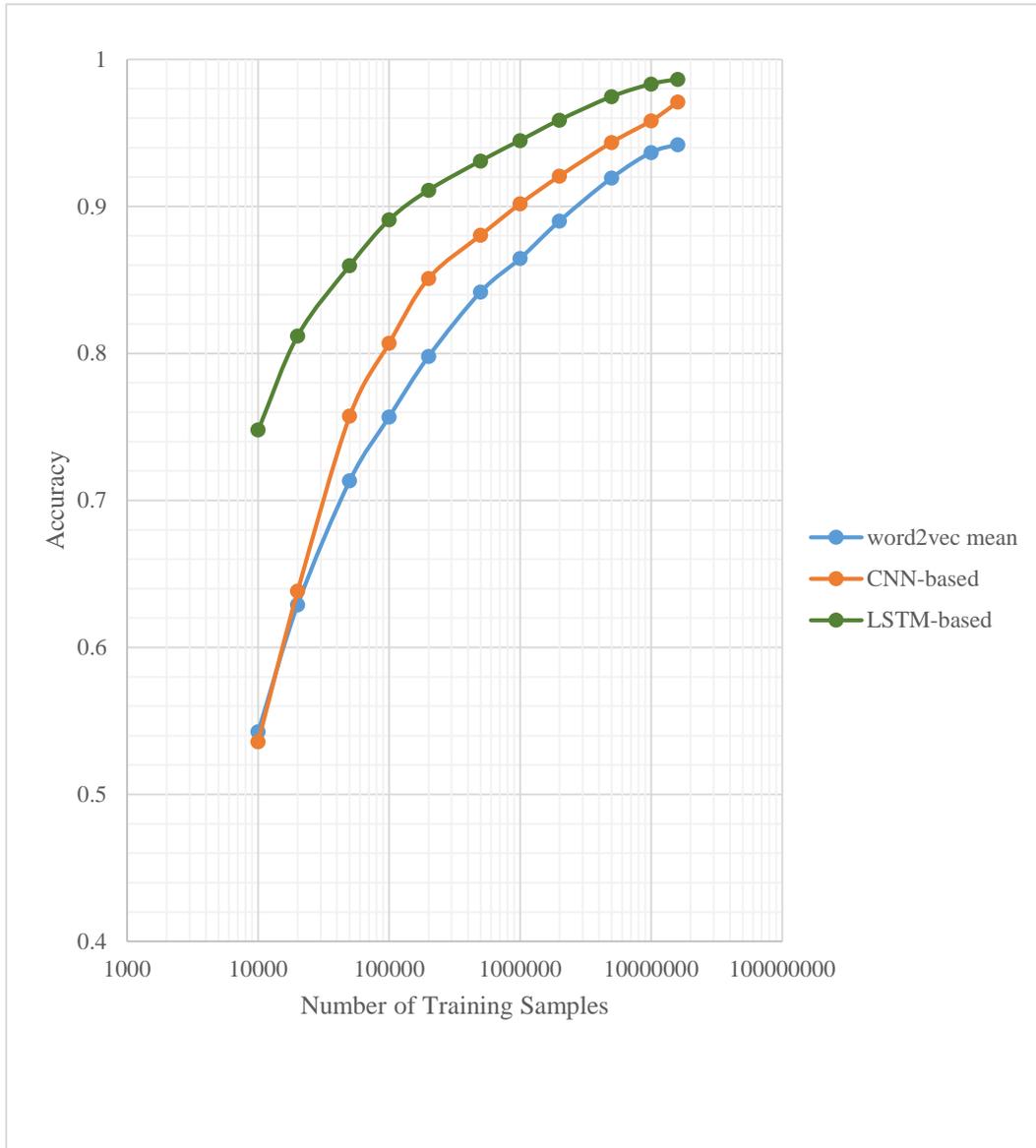

*Figure 3. The effect of the number of the number of training samples on accuracy*

## 6. Conclusion & Future works

We proposed a DNN-based approach for a task which was the combination of semantic textual similarity and text classification tasks in natural language processing field of study. This approach presented an excellent performance in the defined task. We generated the dataset for this task by ourselves with the use of the Divar dataset. We exploited a supervised learning approach and created text pairs which were assigned a score from 0 to 3. This score exhibits the degree of the overlap two texts in each pair have in terms of their categories. We created word embedding vectors and produced text embedding vectors via employing CNN and LSTM in separate experiments. Eventually, a 600-dimensional vector was made by concatenating bit-wise multiplication and absolute difference of two text vectors and fed to one-layer FCNN to extract the probability distribution for the score of each pair. We used a GPU to train our models. Notwithstanding the simplicity of our model, the outcome is stunning.

Although the performances of our systems are exceptional, achieving results near to complete is not far-fetched. Recent improvements in NLP tasks by the use of attention-based networks indicates that these networks probably diminish the error in classification tasks and promises obtaining better results in terms of accuracy.